%% file: document.tex
\documentclass[11pt,a4paper]{article}
\usepackage[hyperref]{acl2021}
\usepackage{times}
\usepackage{latexsym}

\newcommand{\todo}[1]{} 
\renewcommand{\todo}[1]{{\noindent \color{red} \textbf{TODO: } {#1}}}

\usepackage[nolist]{acronym}
\usepackage{graphicx}

\usepackage{microtype}

\aclfinalcopy 

\title{Towards Learning Through Open-Domain Dialog}

\author{	
    Eugénio Ribeiro$^{1,2}$, Ricardo Ribeiro$^{1,3}$, \and David Martins de Matos$^{1,2}$ \\
    INESC-ID Lisboa, Portugal \\
    Instituto Superior Técnico, Universidade de Lisboa, Portugal \\
    Instituto Universitário de Lisboa (ISCTE-IUL), Portugal \\
	\texttt{eugenio.ribeiro@inesc-id.pt} \\
}

\date{}

\begin{document}
\maketitle

\input{sections/acronym}
\input{sections/abstract}

\input{sections/introduction}

\input{sections/overview}

\input{sections/extraction}
\input{sections/learning}
\input{sections/grounding}
\input{sections/conclusions}

\input{sections/acknowledgments}

\bibliographystyle{acl_natbib}
\bibliography{references}


\end{document}

%% file: sections/acronym.tex
\begin{acronym}
    \acro{AI}{Artificial Intelligence}
    \acro{DM}{Dialog Management}
    \acro{EL}{Entity Linking}
    \acro{FCT}{Fundação para a Ciência e a Tecnologia}
    \acro{IL}{Incremental Learning}
    \acro{KB}{Knowledge Base}
    \acro{KBP}{Knowledge Base Population}
    \acro{NLG}{Natural Language Generation}
    \acro{NLP}{Natural Language Processing}
    \acro{NLU}{Natural Language Understanding}
    \acro{OpenIE}{Open-Domain Information Extraction}
    \acro{QA}{Question Answering}
\end{acronym}

%% file: sections/abstract.tex
\begin{abstract}
The development of artificial agents able to learn through dialog without domain restrictions has the potential to allow machines to learn how to perform tasks in a similar manner to humans and change how we relate to them. However, research in this area is practically nonexistent. In this paper, we identify the modifications required for a dialog system to be able to learn from the dialog and propose generic approaches that can be used to implement those modifications. More specifically, we discuss how knowledge can be extracted from the dialog, used to update the agent's semantic network, and grounded in action and observation. This way, we hope to raise awareness for this subject, so that it can become a focus of research in the future. 
\end{abstract}

%% file: sections/introduction.tex
\section{Introduction}
\label{sec:introduction}

The development of artificial agents able to extract knowledge from dialog without domain restrictions and use it to improve their capabilities and adapt to different situations has the potential to change how machines are built to perform certain tasks, as well as how we relate to them. This is in line with the roadmap towards machine intelligence proposed by \citet{Mikolov2016}, which defines communication and learning as two of the fundamental properties that intelligent machines should have. In fact, by combining both with an appropriate body, an artificial agent would be able to naturally communicate with humans and learn in the same way they do, acquiring new knowledge and competences by connecting what is learned through dialog with what is observed in the world. Theoretically, such an agent would be able to learn virtually anything and adapt to new situations, removing the limitations that hand-programmed machines have due to the fact that programmers are not able to predict every possible situation a priori.

An agent able to learn through dialog is a dialog system at its core. However, looking into the research on dialog systems, we can see that it has mainly focused on the development of two different kinds of system. On the one hand, there are task-oriented dialog systems, which focus on the acquisition of the information required to perform a specific task required by the user \citep[e.g.][]{Young2000,Allen2001,Wen2017,Yang2020}. Thus, their conversation capabilities are limited to a single or a small set of domains and restricted by a set of predefined actions that they are able to perform. On the other hand, there are conversational agents that have no restrictions in terms of domain, but only focus on keeping the user engaged in the conversation by generating appropriate responses to user utterances, even if they are not actually able to understand or extract any knowledge from them \citep[e.g.][]{Weizenbaum1966,Lowe2017}. Thus, they are only developed for research or short-term entertainment purposes. Although some recent studies \citep[e.g.][]{Serban2017,Cuayahuitl2019} have explored the use of reinforcement learning approaches to incrementally improve the dialog policy and generate better responses to user utterances, neither task-oriented dialog systems nor conversational agents are able to extract knowledge from the dialog and use it to improve their capabilities. Research on this subject is limited to grounding problems in simple domains \citep[e.g.][]{Yu2017,Thomason2017}, in which the focus is typically not on the linguistic part of the dialog, but rather on the ability to identify observations of certain concepts, or to map concepts into actions. 

The aim of this paper is to encourage further research towards the development of artificial agents able to learn through dialog without domain restrictions. We do that in two ways. First, by identifying the aspects of a generic dialog system that need to be modified in order to allow it to learn through dialog (Section~\ref{sec:background}). Second, by proposing generic approaches that can be applied to achieve the required adaptations. More specifically, we discuss how to extract conceptual knowledge from the dialog (Section~\ref{sec:extraction}), how to use it to update the agent's semantic network (Section~\ref{sec:learning}), and how to ground it in observation and actions (Section~\ref{sec:grounding}).

%% file: sections/overview.tex
\section{Overview}
\label{sec:background}

\begin{figure}
    \centering
    \includegraphics[width=\columnwidth]{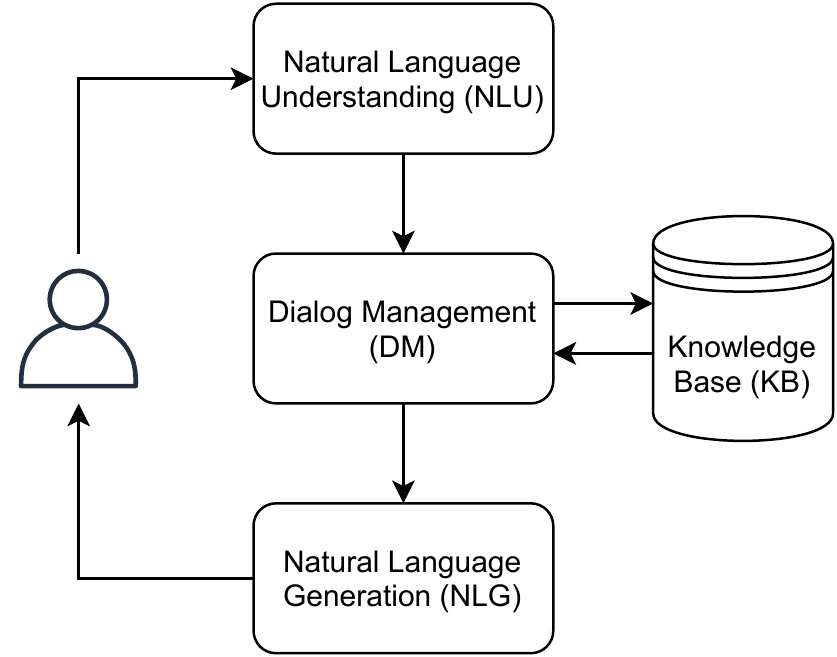}
    \caption{The generic architecture of a dialog system.}
    \label{fig:dialogsystem}
\end{figure}

Dialog systems or conversational agents are programs that interact with their users using some form of natural language. Such systems traverse most areas of \ac{NLP}, because they must be able to interpret user utterances, keep track of the dialog, select the best dialog actions in each context, and generate appropriate responses. Although recent open-domain conversational agents are based on end-to-end models \citep[e.g.][]{Lowe2017,Serban2017} and even task-oriented dialog systems can be developed in an end-to-end fashion that includes API calls \citep[e.g.][]{Byrne2020,Yang2020}, most dialog systems follow the flow depicted in Figure~\ref{fig:dialogsystem}, even if the division into components is implicit. Thus, in the remainder of the paper, we will refer to the components of the modular architecture, as they provide points for anchoring the modifications required to allow a dialog system to learn through dialog. Still, we believe that our considerations can be generalized to end-to-end systems as well. 

Not considering the additional components required to enable spoken interaction, it has long been established that dialog systems can be split into three major components \citep{Young2000,Allen2001}: \ac{NLU}, \ac{DM}, and \ac{NLG}. The first refers to the interpretation of user utterances, by the identifying their domain, intent, and content. The second involves keeping track of the dialog state, communicating with information sources, such as \acp{KB}, and selecting the best dialog actions to take according to the context of the dialog and the system's policy. The last refers to the generation of coherent sentences that transmit the information defined by the selected dialog actions and their parameterization.

The \ac{NLG} ability required for an agent to learn through dialog does not differ from that required in a scenario that does not involve learning. Thus, the modifications mostly concern \ac{NLU} and \ac{DM}. In a task-oriented dialog system, \ac{NLU} consists of the identification of the domain of the utterance, followed by or together with the identification of the domain-specific intent of the user, which includes slots to be filled, defining a frame that maps into a task that can be performed by the system. The \ac{DM} component is then responsible for keeping track of the values of the slots which have already been filled and generating appropriate dialog actions that target the acquisition of the values of the remaining slots. When all the slots are filled, the execution of the corresponding task is triggered, which may involve querying a \ac{KB} to obtain the information required by the user, or calling an external API to execute a command.

On the other hand, in the context of an agent that learns through open-domain dialog, we cannot define sets of slots to be filled, because there are no restrictions in terms of task nor domain. In fact, the concepts and properties that originate those slots may themselves be learned over time through dialog and the domains are fuzzy and inferred from the connections between the concepts known by the agent. Thus, in this scenario, \ac{NLU} has to focus on extracting generic knowledge from the utterances, regarding the existence of concepts and the relations between them. Intent recognition is still important, but from a generic perspective which provides cues for the kinds of knowledge present in each utterance. The identification of such intents may help in the extraction process and allows the \ac{DM} component to find relations between multiple utterances in the dialog.

Additionally, in order for the agent to learn, the \ac{KB} can no longer be just an information source, as it must be updatable as well. In fact, since the structure of the \ac{KB} is itself learned over time based on the concepts and relations extracted from the dialog, it can be generalized as a semantic network \citep{Sowa1991}. The knowledge present in this semantic network can be grounded in action and observation by identifying connections between the concepts and the agent's sensors and actuators. The references to those concepts in the dialog can then be paired with the corresponding observations or actions and used to improve the agent's abilities.

Finally, if we want the agent to be able to guide the dialog towards the acquisition of further knowledge, following an active learning strategy, then the dialog policy must be adapted to also consider prompting for additional information or changing domains. However, it must also consider the context of the dialog and only apply these strategies in appropriate situations.

In the following sections, we propose some approaches for extracting conceptual knowledge from the dialog, use it to update the agent's semantic network, and ground it in action and observation.

%% file: sections/extraction.tex
\section{Knowledge Extraction}
\label{sec:extraction}

In order to learn through open-domain dialog, an agent must be able to extract knowledge from it. \ac{OpenIE} systems \citep{Banko2007} are able to extract relational tuples from text without domain or task restrictions. Thus, they seem appropriate for extracting open-domain knowledge from dialog. However, they usually focus on declarative texts, such as news articles or web data. On the other hand, dialogs include non-declarative utterances, such as questions, and extensive inter-utterance dependencies in the form of answers, corrections, references, among others. Furthermore, \ac{OpenIE} approaches are typically applied to large collections of documents, in which the same information appears multiple times. Thus, they focus on precision instead of recall. On the other hand, in dialogs, each piece of information is transmitted a reduced amount of times, or even only once. Thus, although \ac{OpenIE} systems can serve as base for the extraction of knowledge from dialog, if an existing \ac{OpenIE} system \citep[e.g.][]{Cui2018,Stanovsky2018,Kolluru2020} is applied directly to dialog utterances, it is bound both to extract knowledge that is not factual and to miss important extractions.

The number of missed extractions can be reduced by performing coreference resolution \citep{Sukthanker2020} before applying the \ac{OpenIE} approach. However, this does not solve inter-utterance dependencies based on function nor avoids non-factual extractions. As discussed in Section~\ref{sec:background}, the generic intention behind the utterances can be used to provide cues for these situations. The ISO 24617-2 standard for dialog act annotation \citep{Bunt2017} defines a hierarchy of general-purpose communicative functions that can be automatically identified to a certain extent, even though the amount of annotated dialogs available is reduced \cite{Ribeiro2022}. This hierarchy includes a branch for information-transfer functions that can be used to guide the knowledge extraction process. For instance, the \ac{OpenIE} approach can be applied directly to utterances with an inform function. On the other hand, although different kinds of questions are also able to provide knowledge regarding the existence of concepts or multiple alternatives, their function is to obtain some kind of information. Thus, depending on their kind, questions may include information that is incomplete, uncertain, or even incorrect. To address this problem, questions should be interpreted using approaches similar to those used in the \ac{QA} area \citep{Diefenbach2018}. Additionally, utterances with a responsive function, such as an answer or a correction, can be paired with the utterances they relate to by the \ac{DM} component, leading to the extraction of further knowledge. 

Finally, although the relational tuples extracted by \ac{OpenIE} systems are easily interpretable by humans, in order to be interpreted by a learning agent, they should be extended with information regarding the temporal validity of the relations, and additional knowledge regarding the nature of arguments. In this context, it may be important to include information regarding the generic semantic roles played by the arguments \citep{Petukhova2008}, or even to attempt to identify extractions that evoke similar semantic frames \citep{QasemiZadeh2019}.

%% file: sections/learning.tex
\section{Learning}
\label{sec:learning}

In the previous section, we discussed means to extract knowledge from a dialog. However, in order to learn, the extracted knowledge on its own is not enough. First of all, a learning agent must be able to represent and store what it knows, that is, it must have memory. The conceptual knowledge that can be extracted from dialog consists mainly of concepts, relations between them, and possible restrictions on the scope of their validity. By combining multiple of these extractions, a learning agent can create a semantic network~\cite{Sowa1991} that represents its conceptual memory and, thus, contains the concepts that the agent is aware of. Furthermore, considering that the concepts are identified by their name, this semantic network also defines the vocabulary that the agent can use to talk about its knowledge and which typically maps into a defined semantics known by its conversational partners. Thus, the semantic network is, more specifically, an ontology~\cite{Staab2009}.

Ontologies were originally built by and shared among humans to define a formal context and avoid misunderstandings in communication. However, they can also be used as \acp{KB} queryable by automatic systems, including conversational agents, to obtain knowledge regarding the covered domains. Furthermore, an ontology can also be automatically updated in an incremental fashion, by linking \citep{Shen2014} the concepts and relations referred to in the dialog to those present in the ontology, creating new ones if necessary.

Conflicts may arise during the learning process. Most of these conflicts are easy to identify, because the newly obtained knowledge is incompatible with that present in the \ac{KB}. On the other hand, solving them is not as straightforward. Some conflicts arise due to the existence of ambiguous concepts and can be solved through context disambiguation processes. Other conflicts arise due to misunderstandings or misinformation. Trust- or confidence-based conflict solving strategies can be applied in such situations. However, considering that the agent is in an interactive context, the dialog itself can be used as a tool to solve conflicts, by prompting the conversational partners for the solution.

%% file: sections/grounding.tex
\section{Knowledge Grounding}
\label{sec:grounding}

The processes described in the previous sections allow a learning agent to update its semantic network with conceptual knowledge acquired through dialog. However, that knowledge still has no connection to what is observed in the world. Although that might be enough for an agent whose task is purely dialog-based, it is not for an agent trying to get better at a task involving interaction with its environment through other means. In such scenarios, the knowledge present in the semantic network has to be grounded in action and observation. For that to happen, there has to be a mapping between the agent's sensors and the primitive concepts they are able to observe, as well as between the agent's actuators and the representation of the primitive actions they are able to perform. These mappings can be added directly when a new sensor or actuator is added to the agent, or learned through the dialog. Using these connections, the agent can learn how to perform compound actions, as well as how to identify observations of derivative concepts, based on the compositional and hierarchical relations that it learns through dialog.

In order to improve its ability to recognize concept observations, the agent has to create conceptual models for the corresponding concepts, based on the features provided by the sensors related to those concepts. The models can then be improved over time using an \ac{IL} approach \citep{Gepperth2016} together with the labeled observations obtained by combining references to observable concepts in the dialog with the information provided by the sensors. This is the typical approach used in incremental natural language grounding research \citep[e.g.][]{Cakmak2010,Yu2017,Thomason2017}. However, in those scenarios, there is a predefined set of concepts to be grounded and the whole dialog is focused on that objective. On the other hand, in the context of an agent learning through open-domain dialog, the set of observable concepts is also learned over time, not all utterances refer to observable concepts, and there may be references to past or future observations.

Still in the context of natural language grounding, the agent can also rely on the dialog to adopt active learning strategies, by prompting for the concepts which are being observed at a given time \citep[e.g.][]{Cakmak2010,Thomason2017}, or requesting a demonstration of or feedback on a given action \citep[e.g.][]{Cakmak2012}.

%% file: sections/conclusions.tex
\section{Conclusions}
\label{sec:conclusions}

In this paper, we have raised awareness for the lack of research on artificial agents able to learn through open-domain dialog, identified the modifications required for a dialog system to be able to learn from the dialog, and proposed generic approaches that can be used to implement those modifications. This way, we hope that this subject can become a focus of research in the future. 

%% file: sections/acknowledgments.tex
\section*{Acknowledgments}

Eugénio Ribeiro is supported by a PhD scholarship granted by \ac{FCT}, with reference SFRH/BD/148142/2019. Additionally, this work was supported by Portuguese national funds through \ac{FCT}, with reference UIDB/50021/2020.